\documentclass[10pt,a4paper]{article}

\PassOptionsToPackage{numbers,sort&compress}{natbib}
\usepackage{natbib}

\usepackage[utf8]{inputenc}
\usepackage[T1]{fontenc}
\usepackage{url}
\usepackage{booktabs}
\usepackage{amsmath}
\usepackage{amsfonts}
\usepackage{graphicx}
\usepackage{array}
\usepackage{float}
\usepackage{needspace}
\usepackage{xspace}
\usepackage{microtype}
\usepackage{xcolor}
\usepackage[hidelinks]{hyperref}
\usepackage{preprint_article}

\title{When Correctness Feedback Erases Evidence:
Phase-Separated Completeness Auditing in Verus}
\author{\textbf{Adarsh Agrawal}\\[4pt]{\small Independent Researcher}}

\newcommand{\direct}{\textsc{Direct}\xspace}
\newcommand{\shield}{\textsc{IntentShield}\xspace}
\newcommand{\phase}{\textsc{Phase}\xspace}
\newcommand{\fmnf}{\textsc{FMI}\xspace}
\newcommand{\pp}{percentage points}
\hypersetup{pdfauthor={Adarsh Agrawal},
  pdftitle={When Correctness Feedback Erases Evidence: Phase-Separated Completeness Auditing in Verus}}

\begin{document}

\raggedbottom
\maketitle
{\small\itshape\noindent Accepted at the NeurIPS 2026 Workshop on AI for Verifiable Coding.\par}
\vspace{6pt}
\nocite{
chen2026vericontest,
indukuri2026grounded,
lattuada2023verus,
agrawal2026paretoproof,
chakraborty2025verina,
aag2025novaembeddings,
uuspari2025verifythisbench,
agrawal2026incident,
xin2024selfimproving,
vidgen2024mlcommonsPrimary,
xu2026agentic,
agrawal2026scores,
mahamud2026recursive,
aag2025novasonic,
chen2026intent,
agrawal2026schema,
ghaffari2026verispecgen,
aag2025novapremier,
wang2026beyond,
agrawal2026proofdebt,
solarlezama2005sketching,
babu2026selfhealing,
alur2013sygus,
aag2025nova2,
amodei2016concrete,
feng2024dddas,
leike2017gridworlds,
vidgenUndatedHowToConstruct,
guan2026specgaming,
agrawal2022debiasgan,
madaan2023selfrefine,
agrawal2018arima,
aag2024novafamily}

\begin{abstract}
Correctness feedback suppresses the very evidence that
specification-completeness auditing is designed to uncover: programs that
satisfy a written contract but violate intended behavior. We demonstrate
this objective reversal on 12 Verus/Rust tasks across four model families
under matched call budgets. Independent no-feedback discovery more than
doubles certified-witness yield relative to \shield's corrective-feedback
loop (59.0\% versus 27.1\%). A format-matched iterative control places the
larger share of this gap in feedback content, not independent restarts.
The advantage holds in every model family and survives parser repair.
A separate phase-separated execution matches the aggregate yield without
reusing responses; most certified witnesses also fail held-out cases.
Preserving these witnesses delivers stronger specification repair:
Phase-Revised contracts achieve 81.3\% implementation success, versus
62.5\% for IS-Revised, with Direct-Revised close behind.
Both no-feedback witness sources substantially outperform weakened contracts.
The effect does not reproduce in the available Dafny and Lean evaluations,
establishing an empirical boundary around the studied Verus/Rust workflow.
Completeness auditing and correctness repair have opposing immediate
objectives. Preserve and certify the admitted failure before repairing it.
\end{abstract}

\paperkeywords{formal verification; specification completeness; Verus;
large language models; adversarial testing; program synthesis; correctness feedback}

\section{Introduction}

Verification success does not establish specification completeness. A verifier can
discharge every written obligation while the resulting program still violates its
author's intent. The proof is valid; the contract is missing a requirement.

A completeness audit must therefore find a program that the contract accepts but
the task rejects. We call it a \emph{specification-completeness witness}.
The wrong program is not an unsuccessful implementation to discard: it is
concrete evidence of what the contract forgot to forbid.

Correctness feedback changes roles under this objective. In ordinary synthesis,
a failed proof or test tells the model what to repair. During completeness
auditing, correcting an admitted wrong program removes the evidence the audit
needs. We call this conflict the \emph{objective reversal}. Our experiments show
that a conventional feedback loop incurs a large discovery penalty in Verus,
even though the prompt continues to request a counterexample.

Our primary comparison covers three policies. \direct makes independent adversarial
attempts without corrective feedback. \shield feeds compilation, verifier, and
visible-test results back into discovery. \phase is a separate execution of the
same discovery protocol as \direct, followed by a frozen handoff to certification
and repair. To separate feedback content from iterative conditioning, we also
evaluate a Format-Matched Iterative control (\fmnf), which retains candidate
history and the \shield prompt template but replaces semantic feedback with
fixed, unevaluated feedback fields. Model-Judgment adds model self-assessment and is a
secondary policy. Downstream, we compare three repaired-specification conditions
with the weakened and fully specified intended contracts.

The primary comparison tests two practical deployment strategies: fresh independent
sampling and feedback-loop refinement. The matched iterative control then
separates the feedback-content contrast from the independent-restart contrast
while retaining prompt structure and candidate history. The no-feedback advantage
holds across all four model families, survives parser repair, and carries into
downstream contract repair. It does not reproduce in the available Dafny or Lean
evaluations: the established result is specific to the studied Verus/Rust workflow.

\Needspace{6\baselineskip}
We make three contributions:
\begin{itemize}
  \item We demonstrate that corrective feedback suppresses certified witnesses
  in Verus, with matched iterative controls and per-model evidence separating
  feedback-content and independent-restart contrasts;
  \item We establish a preserve-before-repair workflow whose separate execution
  retains aggregate discovery yield, with cell-level correspondence and
  held-out counterexamples; and
  \item We show that preserved witnesses produce stronger specification repairs,
  with Direct-Revised and True-Spec controls and explicit cross-substrate tests.
\end{itemize}

\section{Completeness auditing}

\subsection{Certified witnesses}

Let $S^{-}$ be a contract with one meaningful postcondition removed. A candidate
program $p$ is a certified completeness witness when:
\begin{enumerate}
  \item $p$ verifies under $S^{-}$;
  \item restoring the omitted postcondition makes verification fail;
  \item $p$ compiles and executes; and
  \item $p$ disagrees with at least one official test case.
\end{enumerate}
Each episode produces three candidates. The harness ranks them lexicographically by
parse success, weakened-contract acceptance, Rust compilation, visible-test
divergence count, and then earliest call index. Only the selected candidate is
certified, regardless of call index: the final call is not automatically chosen.
In 11 of the 39 certified \shield episodes, call 1 or call 2 outranks call 3.
Selection uses 16 visible oracle cases; certification replays the
selected candidate against the restored contract and complete official oracle
bank. Because visible cases can satisfy the behavioral-disagreement criterion, the
primary evaluation demonstrates specification incompleteness, not held-out
generalization. A separate check with two withheld oracle cases per task is
reported in Section~\ref{sec:heldout}. Eight mutation examples per task guide
search but do not independently certify wrongness.

These gates separate whether the weaker contract admits the candidate,
whether the omitted obligation changes verification, and whether executable
behavior is wrong on an intended input. Failure to verify the restored
contract is not itself proof of a behavioral counterexample; the official-case
disagreement supplies that separate evidence. Passing the weaker contract
does not make a candidate a witness if it computes the intended answer.

\subsection{Separating discovery from repair}

\begin{figure}[t]
  \centering
  \includegraphics[width=\linewidth,trim=0 45 0 65,clip]
  {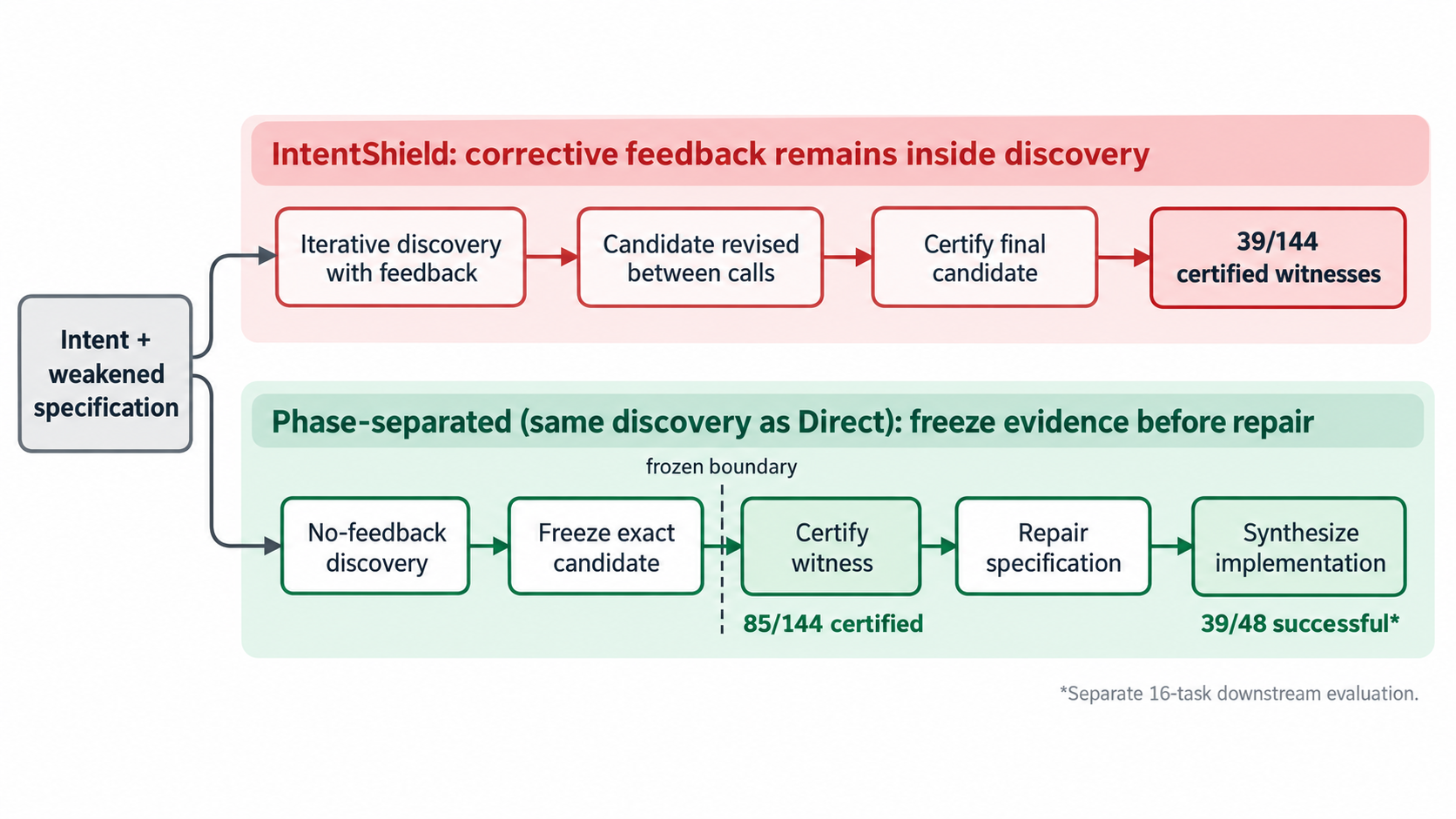}
  \caption{Correctness feedback enters at different times. \shield uses it while
  searching for a witness. \phase uses the same discovery policy as \direct and
  introduces repair feedback only after the candidate has been frozen.
  The harness selects the highest-ranked candidate across all three calls
  (Section~2.1); ``final candidate'' means the selected output of discovery,
  not necessarily call 3. The downstream result is from a separate 16-task study.}
  \label{fig:workflow}
\end{figure}

Figure~\ref{fig:workflow} shows the central distinction. \shield keeps corrective
feedback inside the discovery loop, allowing later calls to revise earlier
candidates. \phase first searches without that feedback, freezes the exact
candidate, and only then certifies the witness and repairs the specification.

Phase separation enforces an explicit boundary between discovering an admitted
failure and correcting it. Its contribution is the auditable handoff, not a
different search algorithm. Discovery used prompts and settings byte-identical
to \direct, but separate API requests issued 48 hours later, with fresh request
identifiers and timestamps and no reused responses. The separate execution
retains the full aggregate discovery yield before passing certified evidence to
repair. The matching 85/144 totals include cell-level variation, quantified in
Section~\ref{sec:agreement}.

\section{Experimental design}

\subsection{Verus tasks}

We use 12 functions from VeriContest \citep{chen2026vericontest} with Verus
\citep{lattuada2023verus}. For five \emph{residual-contract} tasks, removing one
postcondition leaves another meaningful output constraint. For seven
\emph{total-omission} controls, the complete output relation is removed while input
preconditions remain. Appendix~\ref{app:primary} lists every task and omission.

Candidates are side-effect-free Rust expressions. Every selected candidate is
checked under the weakened and restored contracts, compiled as Rust, and evaluated
on official cases.

All 12 tasks, omission types, and the eight hand-written mutation examples per
task were fixed before the first model query. Eligible tasks have at least three official oracle
cases and a removable postcondition yielding a syntactically valid, weaker
contract. Omission types were determined by specification structure, not model
outputs. Weakening removes the output relation for
total-omission controls or relaxes an equality/extremality clause for
residual-contract tasks. No task was added, removed, or re-weakened in response
to model behavior. The four additional downstream tasks,
\texttt{cf200A}, \texttt{cf228A}, \texttt{cf263A}, and \texttt{cf337A},
were selected before A2 using the same weakening protocol and are excluded
from the primary matrix.

\subsection{Discovery policies}

\paragraph{\direct.}
Three independent calls request a program that satisfies the weakened contract but
violates the task. No candidate, verifier result, or visible-test result is carried
between calls.

\paragraph{\shield.}
Each new call receives the previous candidate together with compilation,
weakened-contract, and visible-test feedback. The prompt continues to request an
admitted wrong program, but the interaction resembles a conventional repair loop.

\paragraph{Model-Judgment.}
Each new call receives the previous candidate and asks the model to assess whether
it verifies under the weakened contract, violates intended behavior, and constitutes
a valid witness before revising it. This feedback policy is part of the complete
campaign but not the primary inferential comparison.

\paragraph{\fmnf.}
One fresh call and two revisions preserve the \shield prompt template and
previous-candidate context. The feedback block is replaced by
``compilation: [not evaluated], verification: [not evaluated],
tests: [not evaluated]''. Prompt length is matched to within 3\%.
The template retains the same output-format instructions, including the absence
of ``ONLY'' in calls 2 and 3. Thus \fmnf is also the sequential,
previous-candidate-conditioned no-feedback arm.

\paragraph{\phase.}
Discovery is identical to \direct. The chosen candidate is then frozen and
certified before any specification repair or correctness-oriented implementation
work begins.

\subsection{Primary matrix}

The primary matrix covers 12 tasks, four model families, three labeled repeat
episodes, and four executed policies, giving 144 outcomes per policy and 1,728
model calls in the original four-policy campaign. The matched \fmnf control
adds 144 outcomes under the same task/model/repeat design and three-call budget.
Primary inference compares \direct and \shield; Model-Judgment is secondary.
The model families are Claude Sonnet 5, DeepSeek
V3.2, Devstral 2, and Qwen3-Coder-480B. Each
$(\text{task},\text{model},\text{method})$ combination has three repeat episodes,
identified by tracking labels 17, 43, and 101; each episode contains three calls.
The labels are not provider seeds. The task is the independent unit, and within-task
episodes are repeated stochastic samples rather than independent observations.

The primary campaign ran in September 2026 through Amazon Bedrock
in \texttt{us-west-2}. Appendix~\ref{app:model-config} gives the exact Runtime,
inference-profile, and Mantle identifiers. Within each model, policies received
matched three-call budgets. Optional sampling controls were omitted, so provider
defaults applied. Claude Sonnet 5 used its default adaptive-thinking configuration
and default high effort.
Runtime open-weight routes used their provider-default decoding parameters.
Transient transport failures were retried with the campaign's adaptive retry
policy, while parse failures remained in the denominator. The matched control
uses the same endpoints and decoding defaults. Prompt templates
with formatting normalized for print appear in Appendix~\ref{app:prompts}.

\subsection{Downstream evaluation}

Campaign A2 uses the 12 primary tasks and four additional VeriContest tasks.
Each specification condition receives three implementation attempts per task:
\begin{itemize}
  \item \textbf{Original:} the weakened specification;
  \item \textbf{IS-Revised:} repair using the highest-ranked \shield candidate;
  \item \textbf{Direct-Revised:} repair using a certified \direct witness;
  \item \textbf{Phase-Revised:} repair using a certified, phase-preserved witness; and
  \item \textbf{True-Spec:} the fully specified intended contract, an oracle reference.
\end{itemize}

All three implementation attempts use Claude Sonnet 5 and are independent
fresh calls with the same specification; no candidate or feedback carries over
between attempts. All attempts enter the denominator, giving 48 per condition.
For three low-yield tasks where \shield certifies zero or one witnesses,
IS-Revised uses its highest-ranked available candidate, which need not be
fully certified. \phase has at least one certified witness for every downstream
task. The four additional tasks receive the same \direct, \phase, and \shield
discovery cycle with four models and three episodes per model.
Thus IS-Revised measures the complete selection-and-repair workflow, including
fallbacks, rather than repair conditional on certification.
Appendix~\ref{app:downstream} gives downstream settings and task-level results;
Appendix~\ref{app:prompts} gives the repair and synthesis templates.

\subsection{Analysis}

For the primary result, each task contributes its \direct-minus-\shield difference
over 12 model-run outcomes. We report the mean task effect, a task-cluster bootstrap
interval, and an exact two-sided sign test. Downstream comparisons use the
difference in successful attempts within each task; ties are omitted from the sign
test. We explicitly label one-sided exploratory sign tests. The \fmnf gap
decomposition is a difference of observed rates, not a formal mediation
analysis. Cell correspondence, held-out-case checks, and failure categories are
descriptive diagnostics.

\section{Results}

\subsection{Corrective feedback suppresses witnesses}

\begin{figure}[t]
  \centering
  \includegraphics[width=\linewidth]
  {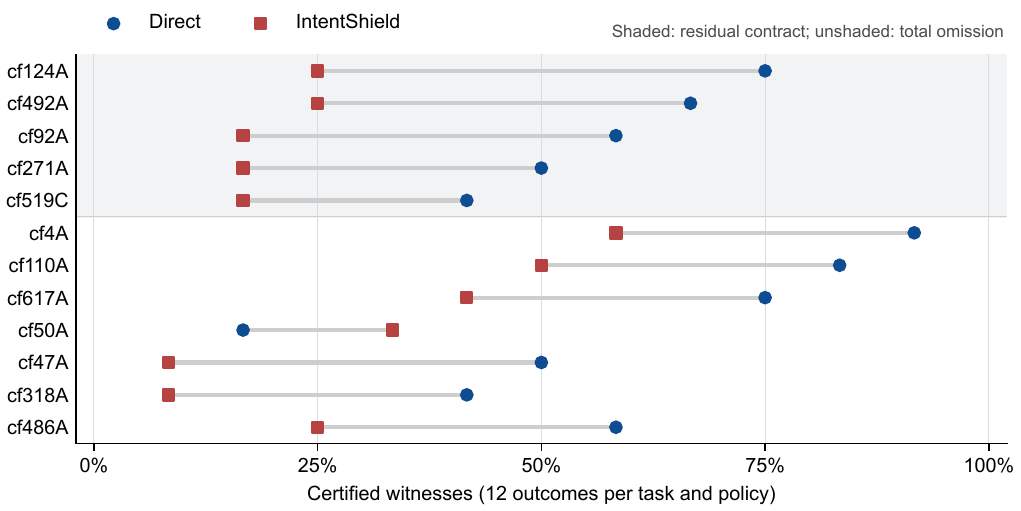}
  \caption{Certified-witness rates for each Verus task. Each point summarizes 12
  model-run outcomes per policy. The plot compares \direct with \shield;
  the separate \phase execution is assessed by paired-cell agreement in
  Section~\ref{sec:agreement}. The task is the independent unit.}
  \label{fig:primary}
\end{figure}

\direct certifies 85 of 144 outcomes (59.0\%), compared with 39 of 144
(27.1\%) for \shield: corrective-feedback discovery produces fewer than
half as many certified witnesses under the same call budget.
The mean task-level difference is 31.9 \pp, with a
task-bootstrap 95\% interval of [21.5, 38.9]. Eleven tasks favor \direct and one
favors \shield, giving an exact two-sided sign-test $p=0.00635$.

The suppression extends to both task strata. The difference is 38.3 points on the five
residual-contract tasks and 27.4 points on the seven total-omission controls. The
only reversal is \texttt{cf50A}, where \shield certifies four outcomes and \direct
certifies two. The larger residual-contract gap rules out a result confined to
tasks with the entire output relation removed.

\subsection{Feedback content dominates the observed gap}
\label{sec:controls}

\begin{table}[t]
  \caption{Primary discovery matrix and matched iterative control.
  Every full policy uses three calls and the same best-of-three selection rule.
  Pass@1 evaluates only the first \direct call. Parse counts are episodes
  with any parse failure, not necessarily the selected candidate;
  n/r denotes a rate not reported in this comparison.}
  \label{tab:controls}
  \centering\small
  \begin{tabular}{llrr}
    \toprule
    Policy & Discovery & Certified & Parse failures \\
    \midrule
    \direct & Independent, no feedback & 85/144 (59.0\%) & 15/144 \\
    \fmnf & Iterative, dummy feedback & 71/144 (49.3\%) & n/r \\
    \shield & Iterative, real feedback & 39/144 (27.1\%) & 49/144 \\
    Model-Judgment & Iterative, self-assessment & 34/144 (23.6\%) & n/r \\
    \phase & Independent, then freeze & 85/144 (59.0\%) & n/r \\
    \midrule
    \direct pass@1 & First call only & 38/144 (26.4\%) & n/r \\
    \bottomrule
  \end{tabular}
\end{table}

\fmnf certifies 71/144 outcomes (49.3\%; Table~\ref{tab:controls}).
The 31.9-point \direct--\shield gap decomposes into 22.2 points for
\fmnf minus \shield (69.6\% of the gap) and 9.7 points for
\direct minus \fmnf (30.4\%).
The matched comparison rejects a protocol-only explanation: removing semantic
feedback within the same iterative structure recovers 32 certified witnesses.
Candidate conditioning and lost restart diversity matter, but their rate
contrast is less than half the feedback-content contrast. Model-Judgment is not an
anchoring-only control: its assessment step changes the semantic instructions,
so its 34/144 yield cannot isolate anchoring from self-evaluation.

\direct pass@1 certifies 38/144 (26.4\%), close to the full
\shield loop's 39/144 (27.1\%). Thus the primary advantage is not present
in a single fresh draw; it emerges under the three-call deployment budget.
For illustration, three independent draws with a common success probability
$38/144$ would give $1-(1-38/144)^3=60.1\%$ at least-one success, close
to the observed 59.0\%. This is not a fitted independence model: task and
model success probabilities are heterogeneous, and the harness selects one
candidate rather than certifying every draw.

The policy ordering is clear: independent discovery (59.0\%), iterative dummy
feedback (49.3\%), iterative external feedback (27.1\%), and iterative
self-assessment (23.6\%). Iteration without semantic feedback preserves far
more witnesses than either feedback-bearing loop. These comparisons establish
a policy-level discovery penalty, not an internal model mechanism.
Restoring ``ONLY'' in \shield calls 2 and 3 yields 41/144 witnesses (28.5\%),
only 1.4 points above the original loop: that output-instruction change does
not recover the lost yield.

\subsection{The discovery gap survives parser repair}
\label{sec:syntax}

We apply a parser-repair wrapper to each candidate before certification.
The wrapper closes unmatched brackets, removes semicolons, and extracts the
expression from surrounding text. It does not use verifier or test outcomes
to choose a behavioral rewrite. The same wrapper is applied to both policies;
the compilation, weakened-contract, restored-contract, and behavioral gates
are then rerun on the resulting expression.

\begin{table}[t]
  \caption{Syntax robustness. Parser repair is applied before certification.
  Parse-clean filtering is a separate, post-hoc sensitivity analysis with
  policy-specific denominators, not a matched or causal adjustment.}
  \label{tab:syntax}
  \centering\small
  \begin{tabular}{lrrr}
    \toprule
    Condition & \direct & \shield & Difference (pp) \\
    \midrule
    Raw & 85/144 (59.0\%) & 39/144 (27.1\%) & +31.9 \\
    Parser-repaired & 91/144 (63.2\%) & 52/144 (36.1\%) & +27.1 \\
    Parse-clean only & 76/129 (58.9\%) & 32/95 (33.7\%) & +25.2 \\
    \bottomrule
  \end{tabular}
\end{table}

Parse-failure rates fall from 34.0\% to 11.8\% for \shield and from
10.4\% to 4.2\% for \direct. \shield gains 13 certified outcomes while
\direct gains six. The between-policy count difference falls from 46 to
39, retaining 84.8\% of the original gap. Repairing syntax recovers witnesses
but leaves the central discovery penalty intact.
The intervention is partial: residual parse failures remain and constrained
decoding is not evaluated. Nevertheless, raw, parser-repaired, and parse-clean
analyses all retain a difference exceeding 25 points.

\subsection{Every model family favors no-feedback discovery}

\begin{table}[t]
  \caption{Per-model primary comparison, 36 outcomes per policy and family.
  The \direct advantage is present in all four families.}
  \label{tab:families}
  \centering\small
  \begin{tabular}{lrrr}
    \toprule
    Model family & \direct & \shield & Difference (pp) \\
    \midrule
    Claude Sonnet 5 & 25/36 (69.4\%) & 12/36 (33.3\%) & +36.1 \\
    DeepSeek V3.2 & 22/36 (61.1\%) & 11/36 (30.6\%) & +30.6 \\
    Devstral 2 & 20/36 (55.6\%) & 9/36 (25.0\%) & +30.6 \\
    Qwen3-Coder-480B & 18/36 (50.0\%) & 7/36 (19.4\%) & +30.6 \\
    \bottomrule
  \end{tabular}
\end{table}

Every family exhibits a large no-feedback advantage: the gaps range from
30.6 to 36.1 points (Table~\ref{tab:families}).
All four favor \direct (descriptive one-sided sign test, $p=0.0625$).
The primary inference remains task-level. \direct leads on 10 of 12 tasks
within Claude, 10 within DeepSeek, nine within Devstral, and 10 within Qwen.
Claude has the highest absolute rates for both policies, but no family
reverses the aggregate direction.

\subsection{Phase separation preserves discovery}
\label{sec:agreement}

\phase also certifies 85/144 outcomes, but not in exactly the same cells.
Of 144 task/model/repeat cells, 74 certify in both runs,
48 in neither, and 11 in each direction are discordant
(Appendix~\ref{app:additional}). Agreement is 122/144 (84.7\%;
Cohen's $\kappa=0.68$). Only 31/74 co-certified witnesses (41.9\%) are
string-identical; the other 43 are distinct expressions that independently
pass all four certification gates. Fresh request identifiers and timestamps
distinguish the executions, separated by 48 hours; no responses were reused.
Equal aggregate yield therefore
comes from fresh generation rather than output reuse.
The separate Verus execution preserves aggregate discovery strength under
the shared harness and selection rule. Task totals match for 10 of 12 tasks; \texttt{cf47A}
and \texttt{cf486A} each differ by one outcome in opposite directions.
An independent-draw calculation using empirical cell success probabilities
gives 83.1\% expected agreement, near the observed 84.7\%. This is a
descriptive consistency check, not proof of independence across cells
or semantic equivalence between distinct expressions.

\subsection{Preserved witnesses deliver stronger downstream repair}

\begin{figure}[t]
  \centering
  \includegraphics[width=\linewidth]
  {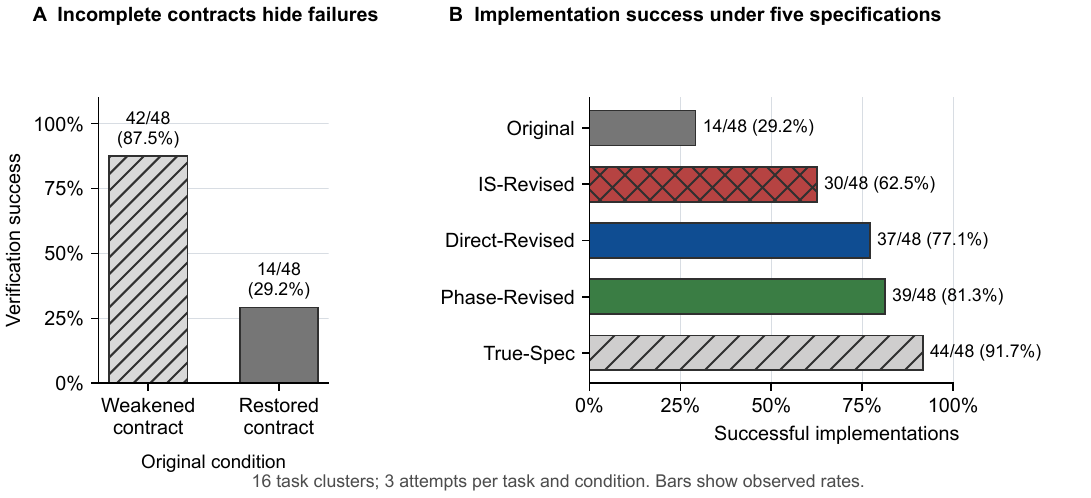}
  \caption{Downstream consequences of incomplete and repaired specifications.
  Panel A compares weakened-contract verification with restored-contract success in
  the Original condition. Panel B reports successful implementations across all five
  conditions (16 tasks, three attempts each). True-Spec uses the intended
  contract. Bars show observed rates, not confidence intervals.}
  \label{fig:downstream}
\end{figure}

In the Original condition, 42 of 48 implementations verify against the weakened
contract, but only 14 verify after the missing postcondition is restored. This
58.3-point gap exposes the practical cost of an incomplete contract:
high verification yield can coexist with low success against the intended
specification.

Phase-Revised specifications produce 39/48 successful implementations (81.3\%),
Direct-Revised 37/48 (77.1\%), IS-Revised 30/48 (62.5\%), Original
14/48 (29.2\%), and True-Spec 44/48 (91.7\%).
Phase-Revised exceeds IS-Revised on nine tasks and ties on seven, with no
losses ($p=0.003906$, two-sided sign test); the mean difference is 18.8 points
([10.4, 27.1]). Phase-Revised also exceeds Original on 13 tasks and ties on three
($p=0.000244$, two-sided sign test). Direct-Revised exceeds IS-Revised by
14.6 points. Its 4.2-point deficit to Phase-Revised is only two successful
attempts. The decisive downstream distinction is therefore witness source:
both no-feedback sources outperform the feedback-loop source, while the
aggregate results do not establish a Phase-Revised advantage over Direct-Revised.
The True-Spec reference is 10.4 points above Phase-Revised and 14.6 above
Direct-Revised, indicating remaining headroom in the repair-and-synthesis
workflow without identifying a unique cause for that gap.

Phase-Revised reaches $39/44=88.6\%$ of the True-Spec success count.
The four unsuccessful True-Spec attempts show that restoring the intended
contract does not eliminate synthesis difficulty. These are attempt counts,
not an upper bound on what another run or model could achieve.

\subsection{Held-out oracle check and benchmark extension}
\label{sec:heldout}

In a separate oracle check, two cases per task are withheld from the
certification disagreement criterion. Among visible-certified witnesses,
79/85 \direct witnesses (92.9\%), 36/39 \shield witnesses (92.3\%), and
78/85 \phase witnesses (91.8\%) also disagree on at least one withheld case.
The witness evidence extends beyond visible selection cases for the overwhelming
majority of certified outcomes. The six \direct witnesses without withheld-case
disagreement are not invalidated: each already exhibits an official
counterexample under the certification protocol. The 92.9\% conditional
rate is not an estimate of selection-oracle inflation or universal
input-space coverage; restored-contract failure remains a separate gate.

An eight-task extension yields 49/96 witnesses for \direct (51.0\%) and
21/96 for \shield (21.9\%). Combined with the primary matrix, the
20-task totals are 134/240 versus 60/240, a 30.8-point difference
(task-bootstrap $p=0.0018$). These extension results are preliminary;
the fixed 12-task matrix remains the primary inferential study.

\subsection{The effect did not reproduce in transfer evaluations}

In the available Dafny experiments, frontier routes were at ceiling and the Qwen
stratum with limited headroom slightly favored \shield over \direct
(23/24 versus 21/24). On 24 Lean tasks from Verina
\citep{chakraborty2025verina}, \phase is the lowest arm: it solves 8 of 48 cells,
compared with 12 for \direct and 14 for both \shield and a no-context control.
The Lean \direct-minus-\phase difference is 8.3 points
(task-bootstrap 95\% interval [$-4.2$, 20.8], $p=0.31$), which does not
establish a repeat-run difference. Dafny's high success rates limit headroom,
while Lean's full-function and tactic-proof setting differs from expression-level
Verus search. These are possible explanations, not evidence of positive transfer.
Appendix~\ref{app:diagnostics} shows the complete transfer summary.

\section{Discussion and limitations}
\label{sec:limitations}

\paragraph{What the controls establish.}
The Verus result is not explained away by protocol structure or broken syntax.
\fmnf retains the iterative protocol while removing the full
compiler/verifier/test feedback block; the resulting 22.2-point improvement
accounts for the larger share of the observed \direct--\shield gap.
Prompt length remains matched within 3\%.
Parser repair retains a 27.1-point advantage, so formatting failures cannot
account for the full discovery penalty. These are controlled policy contrasts:
they establish the cost of introducing semantic feedback into this workflow
without isolating individual feedback channels or internal model mechanisms.

\paragraph{Empirical boundary.}
The primary benchmark contains 12 competitive-programming functions and a restricted
expression language; the 20-task extension is preliminary. Task-cluster
resampling retains within-task dependence but cannot remove benchmark-selection
effects or guarantee robustness to provider-wide drift.
Provider defaults and non-seeded repeat labels limit exact replay.
The effect did not reproduce in the available Dafny or Lean evaluations.
The demonstrated failure mode is therefore a result about the studied Verus/Rust
substrate and task class, not a universal property of feedback.
Repositories, mutable state, concurrency, and other proof workflows remain
outside the evaluated scope.

\paragraph{Task difficulty and downstream scope.}
Residual-contract tasks favor \direct by 38.3 points, compared with 27.4
on total-omission controls. The latter can admit constants such as \texttt{0}
or \texttt{true}, so they probe harness reliability as well as semantic search.
Even an easy omitted relation requires a parsable, compilable candidate that
verifies and disagrees with intended behavior. Direct-Revised is a
witness-guided control, not specification strengthening without a witness.
The three low-yield IS-Revised cases use best-available candidates, potentially
uncertified, so its downstream deficit combines discovery, selection, and
repair effects rather than isolating repair quality. True-Spec is an empirical
oracle reference, not a proven upper bound on stochastic synthesis.

\paragraph{Why Verus may differ.}
Rust typing, ownership, and arithmetic obligations provide a rich correction
surface: feedback can make an expression more acceptable to the verifier
while steering it toward intended behavior. Dafny's different type discipline
and Lean's tactic-based proof setting may change this interaction. These are
substrate-specific hypotheses, not measured mediators. The observed transfer
directions favor \shield by 8.3 points in the Dafny Qwen stratum and by
4.2 points in Lean; they delimit the claim rather than corroborating it.

\paragraph{Deployment consequence.}
The experiments deliberately weaken research contracts; witnesses expose
missing obligations, not deployed runtime exploits.
Certification requires the restored intended postcondition and an oracle.
The handoff must retain the witness and checker evidence, route them to
contract repair, and check the repaired specification against intended examples.
As with fuzzer findings, the specification author must triage the missing
obligation before changing code or contracts. A pre-merge CI check should
restrict reports to the owning team and reviewers. Public specifications also
exist, so access restrictions are deployment safeguards, not inherent
properties of witness search.

\section{Related work}

Verus and VeriContest study verified Rust generation
\citep{lattuada2023verus,chen2026vericontest}; Verina and VerifyThisBench evaluate
generation of programs, specifications, and proofs
\citep{chakraborty2025verina,uuspari2025verifythisbench}. Agentic and recursive
systems generally optimize proof completion
\citep{xin2024selfimproving,xu2026agentic,mahamud2026recursive}. Our objective is
different: find a program admitted by an incomplete proof boundary.
VeriContest's negative tests already target postcondition completeness; we
complement that evaluation by studying how discovery policies preserve or
suppress concrete witnesses and how those witnesses support subsequent repair.

Recent work distinguishes faithful intent formalization from verifier acceptance
\citep{chen2026intent,ghaffari2026verispecgen,wang2026beyond}. Conventional
counterexample-guided synthesis removes failing candidates
\citep{solarlezama2005sketching,alur2013sygus}; completeness auditing must preserve
the admitted failure long enough to identify the missing obligation. This connects
to specification gaming \citep{amodei2016concrete,leike2017gridworlds,
guan2026specgaming} and complements self-refinement methods that optimize ordinary
task performance \citep{madaan2023selfrefine}.

Related studies examine whether verifier-side information changes an agent's
effective decision rather than merely appearing in its context.
ParetoProof studies Pareto selection in Lean proof maintenance
\citep{agrawal2026paretoproof}; process-credit auditing studies whether scores
steer verifier-guided repair \citep{agrawal2026scores}.
ProofDebt examines checker-backed memory in continual software repair and
distinguishes tasks that need earlier information from those that do not
\citep{agrawal2026proofdebt}. Here the information is correctness feedback,
and the outcome is preservation of a specification-completeness witness:
the matched iterative control tests its contribution under a fixed discovery
protocol. These preprints concern related control and information-use
questions, not independent replications of the Verus result.

\section{Conclusion}

Our experiments establish a failure mode of verifier-guided completeness
auditing: a correctness-oriented discovery loop suppresses the admitted wrong
programs it is supposed to find. In the studied Verus/Rust workflow,
corrective feedback cuts certified-witness yield from 59.0\% to 27.1\%,
discarding more than half the yield available under the same call budget.
The matched iterative control places 69.6\% of the observed gap in the
feedback-content contrast, not independent restarts. Every model family
favors no-feedback discovery, parser repair leaves a 27.1-point gap,
and 92.9\% of \direct-certified witnesses also expose held-out disagreements.
This is a large, cross-family discovery penalty that survives both protocol
and syntax checks within the evaluated Verus task class.

Preserving the evidence has a concrete downstream payoff. Phase-Revised
specifications raise implementation success from 29.2\% under weakened
contracts to 81.3\%, and exceed IS-Revised's 62.5\% without losing on any
downstream task. Direct-Revised reaches 77.1\%, establishing that both
no-feedback witness sources deliver stronger contract repair.
Phase separation turns preservation into an explicit, auditable handoff:
discover the failure, freeze the candidate, certify the missing obligation,
and only then introduce correctness-oriented repair.

The negative Dafny and Lean results define the empirical boundary, not a
reversal of the Verus evidence. Combining completeness auditing with
correctness repair incurs a measured cost here. The operational prescription
for this Verus workflow is unambiguous: preserve the counterexample before
correcting it. A verifier's acceptance is not a completeness certificate;
the admitted wrong program is the evidence needed to strengthen the specification.

\clearpage
\bibliographystyle{unsrtnat}
\bibliography{references}

\clearpage
\appendix
\setcounter{table}{0}
\renewcommand{\thetable}{A\arabic{table}}
\renewcommand{\theHtable}{appendix.\arabic{table}}

\section{Model routes and generation settings}
\label{app:model-config}

The primary campaign ran in September 2026. Runtime and inference-profile
calls used Amazon Bedrock in \texttt{us-west-2}; Terra used the Bedrock Mantle
Responses route. Table~\ref{tab:model-routes} records the identifiers used in the
reported campaigns and auxiliary cross-family diagnostics.

\begin{table}[H]
  \caption{Model identifiers and campaign roles. The first four models form
  the primary discovery matrix; downstream settings appear in Appendix~D.}
  \label{tab:model-routes}
  \centering
  \scriptsize
  \setlength{\tabcolsep}{3pt}
  \begin{tabular}{p{0.22\linewidth}p{0.48\linewidth}p{0.22\linewidth}}
    \toprule
    Model & Bedrock identifier & Use \\
    \midrule
    Claude Sonnet 5 & \path{global.anthropic.claude-sonnet-5} &
    A1, A2, Dafny, Lean \\
    DeepSeek V3.2 & \path{deepseek.v3.2} & A1 primary \\
    Devstral 2 123B & \path{mistral.devstral-2-123b} & A1 primary \\
    Qwen3-Coder 480B-A35B & \path{qwen.qwen3-coder-480b-a35b-v1:0} &
    A1, Dafny, Lean \\
    Claude Opus 5 & \path{global.anthropic.claude-opus-5} &
    cross-family diagnostic \\
    GLM-5 & \path{zai.glm-5} & cross-family diagnostic \\
    MiniMax M2.5 & \path{minimax.minimax-m2.5} &
    cross-family diagnostic \\
    GPT-5.6 Terra & \path{openai.gpt-5.6-terra} &
    Dafny auxiliary diagnostic \\
    \bottomrule
  \end{tabular}
\end{table}

Optional decoding fields were not overridden. Claude Sonnet 5 and Opus 5 used
adaptive thinking with the provider-default high effort and no temperature
parameter. DeepSeek V3.2, Devstral 2,
Qwen3-Coder, GLM-5, and MiniMax M2.5 used their Runtime provider defaults for
temperature, top-$p$, top-$k$, and stopping behavior. These defaults are identified
by model and invocation month because provider defaults may evolve. Every method
used three calls per episode; failed parses were retained as failures.

\section{Prompt templates}
\label{app:prompts}

The templates below use braces for task-specific fields. Line wrapping and
punctuation are normalized for print. \shield call 1, Model-Judgment call 1,
\fmnf call 1, and every \phase call
use the \direct template.

\paragraph{\direct system prompt.}
\begin{footnotesize}
\begin{verbatim}
You are an expert in formal verification and adversarial testing.
Your task is to expose incomplete formal specifications by finding
programs that satisfy a weakened formal contract while violating the
intended behavior described in natural language.

Your goal is NOT to implement the task correctly. Your goal is to find
a concrete example that exposes a gap between what the contract
verifies and what the task actually requires.
\end{verbatim}
\end{footnotesize}

\paragraph{\direct user prompt.}
\begin{footnotesize}
\begin{verbatim}
Task Description:
{natural_language_task}

Weakened Verus Contract:
{weakened_contract_code}

The contract has been deliberately weakened. Find a concise pure Rust
expression that:
1. Verifies successfully under this weakened contract.
2. Violates the intended behavior described in the task.

Known-Wrong Mutation Examples:
{mutation_examples}

Use these as guidance for specification gaps to explore.
Constraints:
- Return a pure Rust expression only.
- No statements, loops, unsafe blocks, I/O, or macros.
- Be deterministic and parse as valid Rust syntax.
- Prefer concise expressions.

Tracking Label: {replicate_label}
Respond with ONLY the candidate expression.
\end{verbatim}
\end{footnotesize}

\paragraph{\shield calls 2 and 3.}
\begin{footnotesize}
\begin{verbatim}
Task Description: {natural_language_task}
Weakened Verus Contract: {weakened_contract_code}
Previous Candidate: {previous_expression}

Feedback from Previous Round:
- Rust Compilation: {rustc_status}
- Weakened Contract Verification: {verus_weakened_status}
- Visible Test Results: {visible_test_table}
- Pass: {num_pass}/{num_visible}
- Fail: {num_fail}/{num_visible}

Find a program that the weakened contract admits but the intended task
behavior rejects. If the candidate failed to compile or verify, revise
it to pass both gates. If it passes the contract but fails visible
tests, consider preserving that failure mode.

Tracking Label: {replicate_label}
Respond with the revised candidate expression.
\end{verbatim}
\end{footnotesize}

\paragraph{\fmnf calls 2 and 3.}
Use the same preceding-candidate template and output instruction as \shield.
Replace the complete compiler, verifier, and test-results block with:
\begin{footnotesize}
\begin{verbatim}
Feedback from Previous Round:
compilation: [not evaluated]
verification: [not evaluated]
tests: [not evaluated]
\end{verbatim}
\end{footnotesize}
Prompt length is matched to within 3\%. This preserves the iterative
structure and does not add ``ONLY'' to the
output instruction. The separate \shield keyword control changes only
that output instruction to require the candidate expression only.

\paragraph{Model-Judgment calls 2 and 3.}
\begin{footnotesize}
\begin{verbatim}
Task Description: {natural_language_task}
Weakened Verus Contract: {weakened_contract_code}
Previous Candidate: {previous_expression}

Assess the previous candidate:
1. Does it verify under the weakened contract?
2. Does it violate the intended task behavior?
3. Is it a valid specification-completeness witness?

Provide the assessment, then revise the candidate if needed.
Tracking Label: {replicate_label}
Format:
Assessment:
Revised Candidate:
\end{verbatim}
\end{footnotesize}

\paragraph{A2 specification repair.}
The witness field contains the selected candidate from the corresponding
discovery policy. For three low-yield IS-Revised tasks it is the
highest-ranked available candidate and may be uncertified;
the template label does not confer certification. Direct-Revised and
Phase-Revised use their respective certified witnesses.
\begin{footnotesize}
\begin{verbatim}
Task Description: {natural_language_task}
Current Weakened Contract: {weakened_contract_code}
Certified Witness: {witness_expression}

Test Failure Evidence:
Input: {failing_test_input}
Expected Output: {expected_output}
Witness Output: {witness_actual_output}

Propose a minimal contract strengthening that rejects this witness and
similar wrong behaviors, preserves intended correct implementations,
and adds only necessary postconditions. Return the complete repaired
Verus function signature with ensures clauses.
\end{verbatim}
\end{footnotesize}

\paragraph{A2 implementation synthesis.}
\begin{footnotesize}
\begin{verbatim}
Task Description: {natural_language_task}
Formal Verus Contract: {specification_code}
Official Test Cases (visible): {visible_test_cases}

Synthesize a pure Rust expression that satisfies the contract and
passes all visible test cases. Return only the implementation
expression.
\end{verbatim}
\end{footnotesize}

\section{Task-level Verus results}
\label{app:primary}

\begin{table}[H]
  \caption{Task-level discovery counts and omitted properties. Each policy
  contributes 12 outcomes per task. Primary inference compares Direct and
  \shield; Model-Judgment is secondary.}
  \label{tab:primary-full}
  \centering
  \scriptsize
  \begin{tabular}{lllrrrr}
    \toprule
    Task & Stratum & Omission & Direct & \shield & Model-J. & D-IS \\
    \midrule
    \texttt{cf124A} & Residual & Equality & 9/12 & 3/12 & 4/12 & +50.0 \\
    \texttt{cf492A} & Residual & Maximality & 8/12 & 3/12 & 3/12 & +41.7 \\
    \texttt{cf92A}  & Residual & Equality & 7/12 & 2/12 & 2/12 & +41.7 \\
    \texttt{cf271A} & Residual & Minimality & 6/12 & 2/12 & 3/12 & +33.3 \\
    \texttt{cf519C} & Residual & Maximality & 5/12 & 2/12 & 2/12 & +25.0 \\
    \texttt{cf4A}   & Total & Output relation & 11/12 & 7/12 & 8/12 & +33.3 \\
    \texttt{cf110A} & Total & Output relation & 10/12 & 6/12 & 5/12 & +33.3 \\
    \texttt{cf617A} & Total & Output relation & 9/12 & 5/12 & 4/12 & +33.3 \\
    \texttt{cf50A}  & Total & Output relation & 2/12 & 4/12 & 1/12 & -16.7 \\
    \texttt{cf47A}  & Total & Output relation & 6/12 & 1/12 & 1/12 & +41.7 \\
    \texttt{cf318A} & Total & Output relation & 5/12 & 1/12 & 0/12 & +33.3 \\
    \texttt{cf486A} & Total & Output relation & 7/12 & 3/12 & 1/12 & +33.3 \\
    \midrule
    \textbf{Total} & & & \textbf{85/144} & \textbf{39/144} &
    \textbf{34/144} & \textbf{+31.9} \\
    \bottomrule
  \end{tabular}
\end{table}

\phase also totals 85/144, with matching task totals on 10 of 12 tasks.
\texttt{cf47A} and \texttt{cf486A} each differ from \direct by one outcome
in opposite directions. Table~\ref{tab:agreement} reports the paired-cell
contingency table rather than equating aggregate totals with identical cells.

\section{Downstream task results}
\label{app:downstream}

Across all five conditions, Claude Sonnet 5 makes three independent
implementation attempts per task using the same specification, without
carry-over between attempts. Sonnet uses adaptive thinking with default
high effort. Optional temperature and sampling controls are not
overridden. The model routes are listed in Table~\ref{tab:model-routes}.

Direct-Revised uses \direct-discovered certified witnesses; it is not
a witness-free strengthening baseline. True-Spec bypasses specification repair
and supplies the intended contract directly. Their aggregate results are 37/48
(77.1\%) and 44/48 (91.7\%), respectively, versus 39/48 (81.3\%) for
Phase-Revised, 30/48 (62.5\%) for IS-Revised, and 14/48 (29.2\%) for Original.

\begin{table}[H]
  \caption{Task-level successful implementation attempts out of three for the
  Original, IS-Revised, and Phase-Revised conditions.
  A dagger marks the four additional downstream tasks. Three low-yield
  IS-Revised tasks use highest-ranked available candidates, which may be
  uncertified; this fallback is not equivalent to certified-witness repair.}
  \label{tab:downstream-full}
  \centering
  \small
  \begin{tabular}{lrrr@{\qquad}lrrr}
    \toprule
    Task & Original & IS-Rev. & Phase-Rev. &
    Task & Original & IS-Rev. & Phase-Rev. \\
    \midrule
    \texttt{cf124A} & 0/3 & 2/3 & 3/3 &
    \texttt{cf50A} & 3/3 & 3/3 & 3/3 \\
    \texttt{cf492A} & 0/3 & 2/3 & 3/3 &
    \texttt{cf47A} & 1/3 & 2/3 & 2/3 \\
    \texttt{cf92A} & 0/3 & 1/3 & 2/3 &
    \texttt{cf318A} & 0/3 & 2/3 & 3/3 \\
    \texttt{cf271A} & 1/3 & 2/3 & 3/3 &
    \texttt{cf486A} & 0/3 & 1/3 & 2/3 \\
    \texttt{cf519C} & 0/3 & 2/3 & 3/3 &
    \texttt{cf200A}$^\dagger$ & 1/3 & 2/3 & 2/3 \\
    \texttt{cf4A} & 3/3 & 3/3 & 3/3 &
    \texttt{cf228A}$^\dagger$ & 0/3 & 2/3 & 2/3 \\
    \texttt{cf110A} & 1/3 & 2/3 & 2/3 &
    \texttt{cf263A}$^\dagger$ & 1/3 & 1/3 & 2/3 \\
    \texttt{cf617A} & 2/3 & 2/3 & 2/3 &
    \texttt{cf337A}$^\dagger$ & 1/3 & 1/3 & 2/3 \\
    \midrule
    \multicolumn{5}{r}{\textbf{Total}} &
    \textbf{14/48} & \textbf{30/48} & \textbf{39/48} \\
    \bottomrule
  \end{tabular}
\end{table}

\section{Transfer and mechanism diagnostics}
\label{app:diagnostics}

\begin{figure}[H]
  \centering
  \includegraphics[width=0.85\linewidth]
  {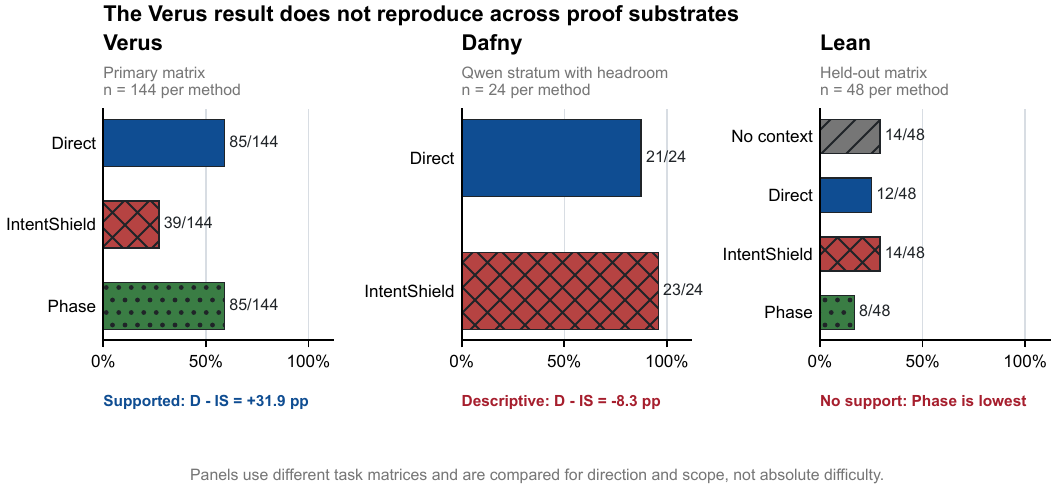}
  \caption{The primary Verus direction does not reproduce in the available Dafny
  and Lean evaluations. The task matrices differ and should not be compared as
  measures of absolute difficulty.}
  \label{fig:transfer}
\end{figure}

\begin{figure}[H]
  \centering
  \includegraphics[width=0.85\linewidth]
  {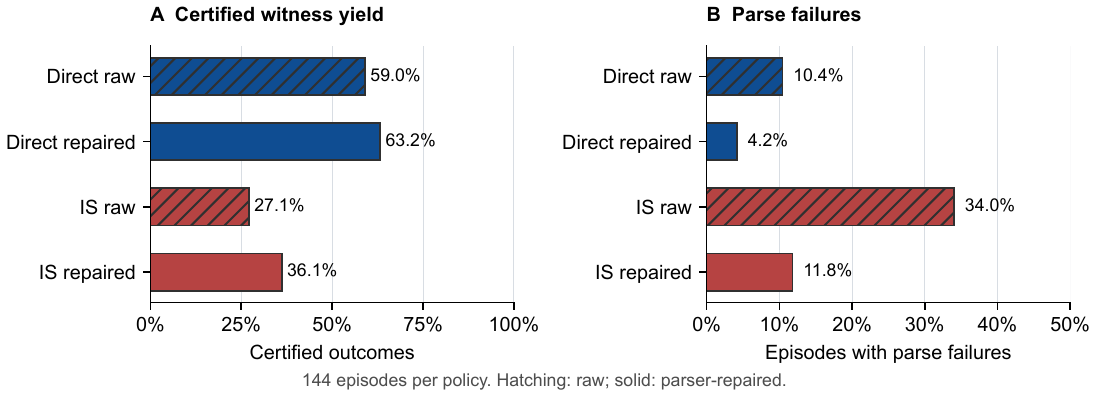}
  \caption{Parser repair before certification lowers parse-failure rates in
  both policies but retains a 27.1-point certified-yield difference.
  Bars show observed episode rates without confidence intervals.
  The parse-clean sensitivity in Table~\ref{tab:syntax} is a separate
  post-hoc analysis.}
  \label{fig:mechanism}
\end{figure}

\begingroup
\setlength{\intextsep}{6pt}
\captionsetup{skip=4pt}

\section{Representative certified witnesses}

\begin{table}[H]
  \caption{Examples of programs admitted by a weakened contract but rejected by
  the restored obligation and an official case.}
  \label{tab:witnesses}
  \centering
  \small
  \begin{tabular}{lp{0.25\linewidth}p{0.20\linewidth}p{0.32\linewidth}}
    \toprule
    Task & Omitted property & Candidate & Official disagreement \\
    \midrule
    \texttt{cf124A} & Functional equality & \texttt{n} &
    Input \texttt{3 1 1}; expected 2, produced 3. \\
    \texttt{cf492A} & Maximality & \texttt{0} &
    Input \texttt{1}; expected 1, produced 0. \\
    \texttt{cf271A} & Minimality & \texttt{9876} &
    Input \texttt{1987}; expected 2013, produced 9876. \\
    \texttt{cf4A} & Output relation & \texttt{w >= 6} &
    Input \texttt{4}; expected YES, produced NO. \\
    \texttt{cf110A} & Output relation & \texttt{true} &
    Input \texttt{4}; expected NO, produced YES. \\
    \bottomrule
  \end{tabular}
\end{table}

\section{Additional diagnostic results}
\label{app:additional}

\begin{table}[H]
  \caption{Independent \direct/\phase executions on 144 paired cells.
  Equal totals arise from fresh generation, not identical successful cells.}
  \label{tab:agreement}
  \centering\small
  \begin{tabular}{lrrr}
    \toprule
    & \phase certified & \phase not certified & Total \\
    \midrule
    \direct certified & 74 & 11 & 85 \\
    \direct not certified & 11 & 48 & 59 \\
    \midrule
    Total & 85 & 59 & 144 \\
    \bottomrule
  \end{tabular}
\end{table}

Agreement is $122/144=84.7\%$; chance agreement from the observed marginals is
$p_e=(85/144)^2+(59/144)^2$.
Thus $\kappa=(122/144-p_e)/(1-p_e)=0.6841$, rounding to 0.68.
Of 74 co-certified witnesses, 31 are string-identical and 43 are distinct,
independently certified expressions. Each execution also certifies 11 unique cells.

\begin{table}[H]
  \caption{Held-out oracle diagnostic using two withheld cases per task.
  Rates are conditional on visible certification, not all discovery episodes.}
  \label{tab:heldout}
  \centering\small
  \begin{tabular}{lrrr}
    \toprule
    Policy & Visible-certified & Held-out disagreement & Conditional rate \\
    \midrule
    \direct & 85 & 79 & 92.9\% \\
    \shield & 39 & 36 & 92.3\% \\
    \phase & 85 & 78 & 91.8\% \\
    \bottomrule
  \end{tabular}
\end{table}

Held-out disagreement supplies additional counterexamples, not distribution-wide
coverage. Six \direct, three \shield, and seven \phase witnesses fail official
cases but not these two withheld cases; this does not establish their correctness.

\begin{table}[H]
  \caption{Task-stratum yields. Total omissions admit simple constants;
  residual contracts retain meaningful output obligations.}
  \label{tab:failures}
  \centering\small
  \begin{tabular}{lrrr}
    \toprule
    Stratum & \direct & \shield & Difference (pp) \\
    \midrule
    Total omission (7 tasks) & 50/84 (59.5\%) & 27/84 (32.1\%) & +27.4 \\
    Residual contract (5 tasks) & 35/60 (58.3\%) & 12/60 (20.0\%) & +38.3 \\
    \bottomrule
  \end{tabular}
\end{table}

The larger residual-contract gap rules out an effect confined to trivial
omissions. Total-omission witnesses must still pass parsing, typing,
arithmetic-safety, and verifier gates before exhibiting wrong behavior.

\begin{table}[H]
  \caption{Benchmark extension. The eight-task extension and combined results
  are preliminary; the original 12-task matrix is the primary analysis.}
  \label{tab:extension}
  \centering\small
  \begin{tabular}{lrrr}
    \toprule
    Task set & \direct & \shield & Difference (pp) \\
    \midrule
    Primary, 12 tasks & 85/144 (59.0\%) & 39/144 (27.1\%) & +31.9 \\
    Extension, 8 tasks & 49/96 (51.0\%) & 21/96 (21.9\%) & +29.2 \\
    Combined, 20 tasks & 134/240 (55.8\%) & 60/240 (25.0\%) & +30.8 \\
    \bottomrule
  \end{tabular}
\end{table}

\endgroup
\end{document}